# Solving Pallet loading Problem with Real-World Constraints

Marko Švaco, Filip Šuligoj, Bojan Šekoranja, Josip Vidaković

University of Zagreb, Faculty of Mechanical Engineering and Naval Architecture

Correspondence: bojan.sekoranja@fsb.hr

https://crta.fsb.hr/en

## ABSTRACT
Efficient cargo packing and transport unit stacking play a vital role in enhancing logistics efficiency and reducing costs in the field of logistics. This article focuses on the challenging problem of loading transport units onto pallets, which belongs to the class of NP-hard problems. We propose a novel method for solving the pallet loading problem using a branch and bound algorithm, where there is a loading order of transport units. The derived algorithm considers only a heuristically favourable subset of possible positions of the transport units, which has a positive effect on computability. Furthermore, it is ensured that the pallet configuration meets real-world constraints, such as the stability of the position of transport units under the influence of transport inertial forces and gravity.

## KEYWORDS

## INTRODUCTION
In the realm of logistics, efficient cargo handling and transportation are paramount to achieving cost-effective and streamlined operations. A key aspect of this process is the optimization of the configuration of transport units on pallets. By maximizing the utilization of available space and ensuring the safety and integrity of the goods, optimization packing practices have garnered significant attention from both industry professionals and academic researchers. Despite its wide range of applications, this distributors pallet loading problem remains one of the most challenging optimization problems to solve, given that it belongs to the class of NP-hard problems.

In this article, we consider the pallet loading problem, where transport units have the rectangular prism shape, the pallet is rectangular, packing is orthogonal, and the loading order of transport units is partially predetermined. In this loading sequence, only skipping of the transport units is allowed. It is assumed that this loading order ensures the structural integrity of the transport units and their contents. Also, we consider vertical and horizontal support of the transport units, in order to ensure position stability. Furthermore, proposed branch and bound algorithm for solving the pallet loading problem uses a heuristic procedure to generate potential positions of transport units on the pallet.

## LITERATURE REVIEW
Since the pallet loading problem belongs to the NP-hard class of problems, literature offers many heuristic solving methods compared to exact methods that find optimal solutions. The following literature review provides an overview of both approaches. There are two main issues with the proposed approaches in the literature. Exact methods are computationally demanding, and can provide optimal solutions only for problems of relatively small size. Heuristic methods, on the other hand, consider only a limited number of potential solutions or overlook practical constraints such as stability and structural integrity of transport units.

Heuristic Methods

The concept of so-called extreme points used for positioning of the transport units is introduced in [1]. The placement of transport units involves sorting with respect to the their dimensions, followed by the application of the first-fit or best-fit heuristic placement method. The approach of space defragmentation together with an extreme point insertion constructive heuristic and a bin-shuffling meta-heuristic strategy is developed in [2], where the unused space between transport units is expanded by moving the boxes along the pallet axes. In [3], the same author presented a greedy look-ahead search algorithm for solving the pallet loading problem, by placing generated blocks of transport units on the pallet. A novel heuristic method based on guided local search is presented in [4]. The search neighborhood is defined by translation of transport units along pallet axes and relocation to another pallet, and it is explored trough transport unit shifting. The algorithm based on stacking layers of transport units is derived in [5]. First, transport units are divided with respect to their height, and layers are then generated accordingly. Then the layers are packed using a branch and bound algorithm for one-dimensional bin packing. Next, instead of the height of transport units for the formation of layers, the base area is used, and a better solution of the two is chosen. In [6], two-level tabu search heuristic is proposed for solving 3D bin packing problem. In first level, the transport units are assigned to the bins, and then in second level, arrangement of units inside bins is determined using interval graph representation of the packing. A hybrid heuristic method combining the variable neighborhood descent structure and the greedy randomized adaptive search procedure is presented in [7].

Exact Methods

In [8], the authors developed a branch and bound based method for placement of transport units at so-called corner points, which are positions generated by already placed transport units. Formulation of this problem in the form of mixed-integer linear programming problem is given in [9], based on relative positions of transport units. This approach was extended in [10], allowing the rotation of transport units placed on pallet, which has a variable height. Furthermore, in [11], stability, load-bearing capacity and fragility constraints were added. This method was further extended in [12], introducing constraints for the shape of the transport unit container, transport unit orientation, and weight distribution. Novel lower bounds based on a relaxation of the mixed integer linear programming problem and valid inequalities are proposed in [13]. These inequalities are inspired by parallel-machine scheduling problems. In [14], algorithm presented in [8] is extended in order to consider all orthogonal packings, by combining original enumerative approach with a new constraint programming approach. A graph-theoretic characterization of the feasible packing arrangements based on relative positions of transport units was proposed in [15], together with a two-level search tree algorithm for solving the multidimensional packing problem.

## METHOD FOR SOLVING PALLET LOADING PROBLEM

In this section, a branch and bound algorithm is derived for solving the pallet loading problem. This type of algorithm can be very efficient because it allows rejection of some solution candidates that do not contain an optimal solution. This is based on the calculation of upper bounds for subsets of candidate solutions. Namely, in each node of the search tree that represents the transport unit placed on the pallet, the upper bound of volume utilization is calculated with respect to the remaining transport units and pallet volume. This upper bound is defined by the following optimization problem:

$$\max \sum_{j \in K} u_j V_j, \quad \sum_{i \in k} u_j V_j \leq V_p, \quad u_j \in (0,1),$$

where $K$ denotes a set of transport units that have not yet been placed on the pallet, $V_j$ represents the volume of transport unit $j$, $V_p$ represents the remaining unused volume of the pallet, and $u_j$ is a binary variable that has the value one if a transport unit $j$ is used to fill the remaining volume of the pallet, otherwise it is zero. Therefore, a computing of the upper bound is done by solving a integer linear programming problem in the form of so-called 0-1 knapsack problem. One method for solving this type of problem is the branch and bound algorithm, which is used in conjunction with the interior point method to find solutions to linear programming problems. The interior-point method is a efficient numerical method that allows finding solutions to linear programming problems in polynomial time. In this case, the branch and bound algorithm searches for the optimal integer solution through an iterative process that involves exploring the search space by maximizing the objective function without the integer variable constraints (interior-point method) and gradually introducing constraints that eliminate the space of non-integer solutions, thus reducing the search space until an integer solution is found. Although in the worst case, the optimization objective function needs to be calculated for all possible combinations of variable values, in practice, it has been shown that this algorithm efficiently finds optimal solutions for problems of limited size.

Let origin of the pallet's coordinate system be located at one of the corners of the pallet, and the coordinate axes be defined by the direction of the pallet's sides. Since we are considering orthogonal packing, the dimensions of the transport units are also defined according to this coordinate system. The unused pallet volume $V_p$ will be defined using discretization of the pallet space. This discretization is obtained by projecting the corner points of the transport units onto the horizontal axes of the pallet coordinate system, and defining the associated two-dimensional mesh grid. Let the array $D_x$ contain the coordinates of the corner points of the loaded transport units with respect to the horizontal axis $x$ of the pallet coordinate system. Also, let $D_y$ be defined in the analogous manner with respect to horizontal axis $y$, and assume data repetitions in these arrays are removed, and that they are sorted in ascending order. Furthermore, let $S$ represent set of transport units loaded on the pallet. Each element $i$ of set $S$ has associated arrays $(x_i, y_i, z_i)$ and $(w_i, d_i, h_i)$, that represent the position coordinates and dimensions of the transport unit $i$, respectively. Each point of the mesh grid defined by the arrays $D_x$ and $D_y$ is assigned a height coordinate. This is the maximum height of transport units whose vertical projection intersect that point. The surface defined by these points together with the maximum permitted pallet height determines the unused pallet volume, since transport units cannot be placed below this envelope, due to the loading order. Therefore, we can define the unused pallet volume $V_p$ by the following expression:

$$V_p = \sum_{j=2}^{n_x} \sum_{k=2}^{n_y} \left(z_p - \max(\{z_i + h_i | i \in S_{j,k}\}, 0)\right)\left(D_x(j) - D_x(j-1)\right)\left(D_y(k) - D_y(k-1)\right),$$

where $z_p$ is the maximum height of the pallet, $n_x$ is the lenght of the array $D_x$, $n_y$ is the lenght of the array $D_y$, and set

$$S_{j,k} = \{i \in S | D_x(j-1) \geq x_i, D_x(j-1) < x_i + w_i, D_y(k-1) \geq y_i, D_y(k-1) < y_i + d_i\}$$

$$\text{for } j = 2, \ldots, n_x, \text{ and } k = 2, \ldots, n_y,$$

defines all transport units whose vertical projection intersect selected point of mesh grid.

Potential positions of transport units on the pallet are defined as so-called extreme points. These points can be obtained by projecting certain corner points of transport units onto the pallet sides or other transport units, in the opposite direction of the pallet coordinate axes, as defined by the following pseudocode:

```
Set Extreme points=();
For each (i ∈ S)
    Set   x_y = (x_i + w_i, max({y_j + d_j|j ∈ S, x_i + w_i < x_j + w_j, y_i ≥ y_j + d_j},0), z_i);
    Set   x_z = (x_i + w_i, y_i, max({z_j + h_j|j ∈ S, x_i + w_i < x_j + w_j, z_i ≥ z_j + h_j},0));
    Set   y_x = (max({x_j + w_j|j ∈ S, y_i + d_i < y_j + d_j, x_i ≥ x_j + w_j},0), y_i + d_i, z_i);
    Set   y_z = (x_i, y_i + d_i, max({z_j + h_j|j ∈ S, y_i + d_i < y_j + d_j, z_i ≥ z_j + h_j},0));
    Set   z_x = (max({x_j + w_j|j ∈ S, z_i + h_i < z_j + h_j, x_i ≥ x_j + w_j},0), y_i, z_i + h_i);
    Set   z_y = (x_i, max({y_j + h_j|j ∈ S, z_i + h_i < z_j + h_j, y_i ≥ y_j + h_j},0), z_i + h_i);
    Set Extreme points =( Extreme points, x_y, x_z, y_x, y_z, z_x, z_y);
End
```

Not all possible potential positions of transport units are covered by extreme points, but in this way we prevent the search tree from becoming too large and reduce the search time, while the generated potential positions are favourable. After generating the potential positions of transport units, it is necessary to determine whether the selected transport unit can be placed in each position. The selected transport unit can be placed in the defined position if the following practical constraints are satisfied:

- The selected transport unit must not overlap with other transport units on the pallet, and must not exceed pallet boundaries.
- The selected transport unit must have vertical support. Vertical support is defined with the percentage of the bottom surface of the transport unit that is supported by other transport units or the pallet. Permissible distance between supporting and supported surfaces is also defined, as transport units have a certain level of deformability.
- The selected transport unit must have horizontal support. Horizontal support is defined as the percentages of the lateral surfaces of the transport unit that are supported by other transport units. This is necessary to ensure a stable position of the transport units during transportation. Horizontal support is considered only for the transport unit surfaces whose normals are in the opposite direction of the pallet coordinate axes. It is assumed that other surfaces are supported by pallet wrap. Permissible distance between supporting and supported surfaces is also defined, as transport units have a certain level of deformability.

This conditions for transport unit placement are checked for two orthogonal orientations of the transport unit. Each feasible combination of position and orientation of the selected transport unit is evaluated based on the relation between the vertical and horizontal support provided by that transport unit and other transport units. In particular, the evaluation will favor those positions and orientations of transport unit in which the support surfaces are in the same plane as the support surfaces of other transport units, taking into account the distance between them. In this way, the creation of defragmented support surfaces is endorsed, thereby increasing the number of possible pallet configurations. A limited number of the best combinations are selected based on these evaluations, so that the search tree includes only those combinations that are heuristically favorable. The following expressions define the sets of transport units whose support surfaces are in the same plane as support surfaces of the selected transport unit at some position and orientation:

$$S_{z,i} = \{j \in S \mid z_i + h_i + p_z \geq z_j + h_j \land z_i + h_i - p_z \leq z_j + h_j\},$$
$$S_{x,i} = \{j \in S \mid x_i + w_i + p_x \geq x_j + w_j \land x_i + w_i - p_x \leq x_j + w_j\},$$

$$S_{y,i} = \{j \in S \mid y_i + d_i + p_y \geq y_j + d_j \wedge y_i + d_i - p_y \leq y_j + d_j\},$$

where the parameters $p_z, p_x$ and $p_y$ denote permitted variation with respect to the supporting plane. Evaluation of the position and orientation of the selected transport unit is defined by the following expression:

$$O_i = \sum_{S_{z,i}} \frac{w_j d_j}{Z_{i,j}} + \sum_{S_{x,i}} \frac{d_j h_j}{X_{i,j}} + \sum_{S_{y,i}} \frac{w_j h_j}{Y_{i,j}},$$

where the distances of the supporting surfaces are defined as:

$$Z_{i,j} = \sqrt{(x_i + \frac{w_i}{2} - x_j - \frac{w_j}{2})^2 + (y_i + \frac{d_i}{2} - y_j - \frac{d_j}{2})^2},$$

$$X_{i,j} = \sqrt{(y_i + \frac{d_i}{2} - y_j - \frac{d_j}{2})^2 + (z_i + \frac{h_i}{2} - z_j - \frac{h_j}{2})^2},$$

$$Y_{i,j} = \sqrt{(x_i + \frac{w_i}{2} - x_j - \frac{w_j}{2})^2 + (z_i + \frac{h_i}{2} - z_j - \frac{h_j}{2})^2}.$$

We will now describe pallet loading problem in the form of tree search problem. Each node in the search tree represents a positioned and oriented transport unit. From each node, the search tree branches to a limited number of nodes. These nodes are generated using the described evaluations of positioned and oriented transport unit and defined constraints. The pallet volume occupied by the loaded transport units determines the lower bound of this optimization problem. After generating branches, thus a new layer of nodes, we compare the lower and upper bounds to determine if we need to search further in these branches to get a better utilization of the pallet volume. This indirect verification of a subset of candidate solutions significantly speeds up the execution of the algorithm. In order to explore the entire search tree, i.e. all combinations of positions and orientations of transport units that are generated by extreme points, satisfy the defined constraints and are considered by defined evaluation, we can use the following algorithm:

```
// Select the first transport unit with respect to the picking order.
// Generate the first node defined by selected transport unit and position (0,0,0).
// Select the first node.
// Select the next transport unit in the picking order.
// Define the maximum number of branches.
// Define the lower bound A equal to the volume of the first transport unit.
Loop L1
     //Define position and orientation candidates for the selected transport unit
     based on extreme points and stability constraints.
     //Evaluate the position and orientation candidates for the selected transport
     unit with respect to the maximum number of branches.
     //Define the lower bound B equal to the sum of the volume of loaded transport
     units and the volume of the selected transport unit (if there are feasible
     positions and orientations).
     If (lower bound B>lower bound A)
          Set lower bound A=lower bound B.
          //Save the generated pallet configuration.
     end
     //Calculate the upper bound with respect to the loaded transport units and
     selected transport unit (if there are feasible positions and orientations).
```

Solving Pallet loading Problem with Real-World Constraints

```
    If (there are feasible positions and orientations for the selected transport
    unit) and (upper bound>lower bound B) and (there are remaining transport units
    with respect to the selected transport unit and the picking order)
            //Generate the next layer of nodes considering the best position and
            orientation candidates of the selected transport unit.
            //Select the first node in the generated layer of nodes.
            //Select the next transport unit in the picking order.
    Elseif (there are no feasible positions and orientations for the selected
    transport unit) and (there are remaining transport units with respect to the
    selected transport unit and the picking order) and (upper bound>lower bound
    B)
            //Select the next transport unit in the picking order.
    Else
            If (there are unexplored node in the layer of the selected node) and
            (there are remaining transport units)
                    //Select next node in layer of the selected node.
                    //Select the next transport unit in the picking order.
            Else
                Loop L2
                    //Select the transport unit defined in the selected node.
                    If (the selected node is not the first node in the search
                    tree)
                            //Backtrack to the last selected node in the
                            previous layer of nodes.
                    Elseif (there are remaining transport units with respect
                    to the selected transport unit and the picking order)

                            //Replace the selected transport unit with the next
                            transport unit in picking order.
                            //Select the transport unit defined in the selected
                            node.
                    Else
                            Return Saved pallet configuration.
                    End
                    If (there are remaining transport units with respect to
                    the selected transport unit and the picking order)
                            //Select the next transport unit with respect to
                            the selected transport unit and the picking order.
                            //Exit loop L2.
                    End
                End
            End
        End
    End
End
```

The experiments were performed on a PC with an Intel Core i7-8750H CPU 2.20 GHz, 16 GB RAM, and Windows 10 Operating System. Algorithms have been implemented in MATLAB. Images below show of examples of pallet configurations obtained by the described algorithm and a time limit of 5 minutes.

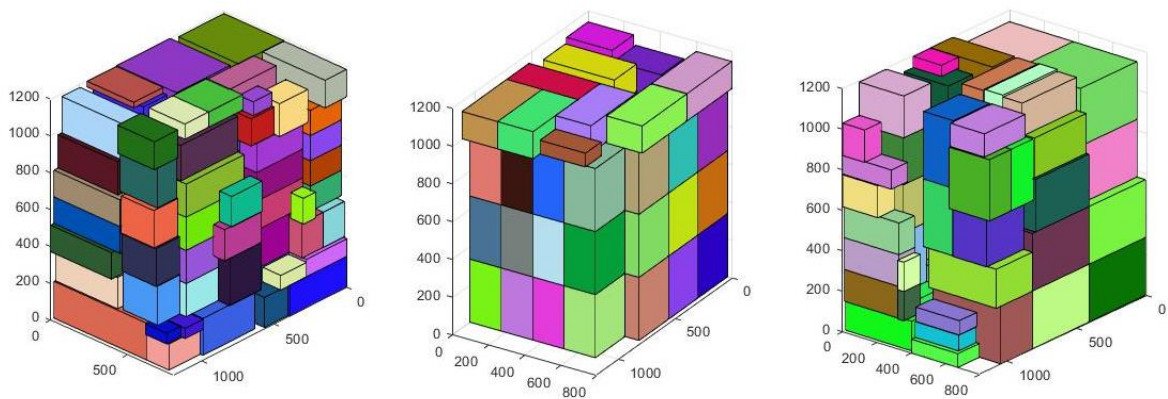

Examples of pallet configuration

# CONCLUSION

The proposed branch and bound algorithm has been shown to be effective for finding pallet configuration solutions that take into account the real-world stability constraints of transport units. The approach of favouring the defragmentation of the collective support surface of transport units seems to be a key element for obtaining applicable solutions. The findings of this study can assist logistics professionals in improving efficiency, reducing costs, and enhancing overall logistics operations. Future research can continue to explore novel algorithms, hybrid approaches, and real-time optimization strategies to further enhance the logistics efficiency and address the complexities of cargo packing in different contexts.

# ABOUT THE PROJECT

# ACKNOWLEDGEMENTS

The authors would like to acknowledge the support of the European Union's European Regional Development Fund which co-financed the project. The content of the publication is the sole responsibility of the Faculty of Mechanical Engineering and Naval Architecture.

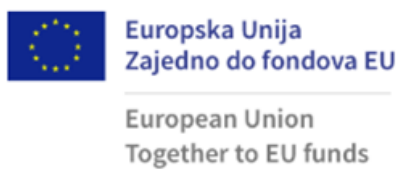
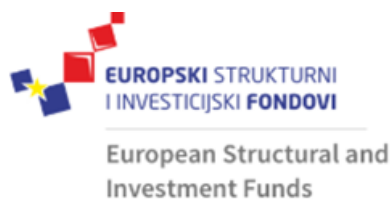